%% file: main.tex
\documentclass[letterpaper]{article} 
\usepackage{aaai2026}  
\usepackage{times}  
\usepackage{helvet}  
\usepackage{courier}  
\usepackage[hyphens]{url}  
\usepackage{graphicx} 
\usepackage{natbib}  
\usepackage{caption} 
\usepackage{algorithm}
\usepackage{algorithmic}
\usepackage{amsmath}
\usepackage{newfloat}
\usepackage{listings}
\DeclareCaptionStyle{ruled}{labelfont=normalfont,labelsep=colon,strut=off} 
\floatstyle{ruled}
\newfloat{listing}{tb}{lst}{}
\floatname{listing}{Listing}
\title{Multi-Agent VLMs Guided Self-Training with PNU Loss for Low-Resource Offensive Content Detection}
\author {
    Han Wang\textsuperscript{\rm 1},
    Deyi Ji\textsuperscript{\rm 2},
    Junyu Lu\textsuperscript{\rm 3},
    Lanyun Zhu\textsuperscript{\rm 4},
    Hailong Zhang\textsuperscript{\rm 2},
    Haiyang Wu\textsuperscript{\rm 2},
    Liqun Liu\textsuperscript{\rm 2}\textsuperscript{\rm *}, \\
    Peng Shu\textsuperscript{\rm 2},
    Roy Ka-Wei Lee\textsuperscript{\rm 1}\textsuperscript{\rm *}
}
\affiliations {
    \textsuperscript{\rm 1}Singapore University of Technology and Design \\
    \textsuperscript{\rm 2}Tencent \\
    \textsuperscript{\rm 3}Dalian University of Technology \\
    \textsuperscript{\rm 4}Nanyang Technological University\\
    \{han\_wang, roy\_lee\}@sutd.edu.sg, 
    \{deyiji, lericzhang, gavinwu, liqunliu, archershu\}@tencent.com, 
     dutljy@mail.dlut.edu.cn,
     lanyun\_zhu@ntu.edu.sg 
}
\usepackage{bibentry}

\begin{document}

\maketitle
\renewcommand{\thefootnote}{\fnsymbol{footnote}} 
\footnotetext[1]{Corresponding authors.}

\begin{abstract}
Accurate detection of offensive content on social media demands high-quality labeled data; however, such data is often scarce due to the low prevalence of offensive instances and the high cost of manual annotation. To address this low-resource challenge, we propose a self-training framework that leverages abundant unlabeled data through collaborative pseudo-labeling. Starting with a lightweight classifier trained on limited labeled data, our method iteratively assigns pseudo-labels to unlabeled instances with the support of Multi-Agent Vision-Language Models (\textsf{MA-VLMs}). Unlabeled data on which the classifier and \textsf{MA-VLMs} agree are designated as the \textit{Agreed-Unknown} set, while conflicting samples form the \textit{Disagreed-Unknown} set. To enhance label reliability, \textsf{MA-VLMs} simulate dual perspectives, moderator and user, capturing both regulatory and subjective viewpoints. The classifier is optimized using a novel Positive-Negative-Unlabeled (PNU) loss, which jointly exploits labeled, \textit{Agreed-Unknown}, and \textit{Disagreed-Unknown} data while mitigating pseudo-label noise. Experiments on benchmark datasets demonstrate that our framework substantially outperforms baselines under limited supervision and approaches the performance of large-scale models. 
\end{abstract}

\begin{links}
    \link{Code}{https://github.com/Social-AI-Studio/MA-VLM.git}
\end{links}

\section{Introduction}

\input{introduction}

\section{Related Works}

\input{related}

\section{Methodology}
\input{model}

\section{Experiment}
\input{experiment}

\section{Ablation Study}

\input{ablation}

\section{Conclusion}

\input{conclusion}

\newpage
\section{Acknowledgements}
\input{Acknowledgement}
\bibliography{aaai2026}




\end{document}

%% file: introduction.tex
Offensive content on social media, including hate speech, misogyny, and harassment, threatens individual well-being, democratic discourse, and public safety. Although major platforms deploy moderation systems, these often fall short in coverage, fairness, and adaptability across languages, cultures, and modalities. As platforms scale globally and harmful content grows more multimodal and diverse, automated detection becomes essential. Yet building robust and equitable systems remains difficult due to the scarcity of high-quality labeled data, especially for minority languages and underrepresented communities. This scarcity is compounded by the labor-intensive nature of annotation, which demands fine-grained understanding of context, sarcasm, and implicit harm. Consequently, many regions and communities remain underprotected by current AI safety efforts.


\begin{figure}[t]
  \centering
  \small
  \includegraphics[width=0.47\textwidth]{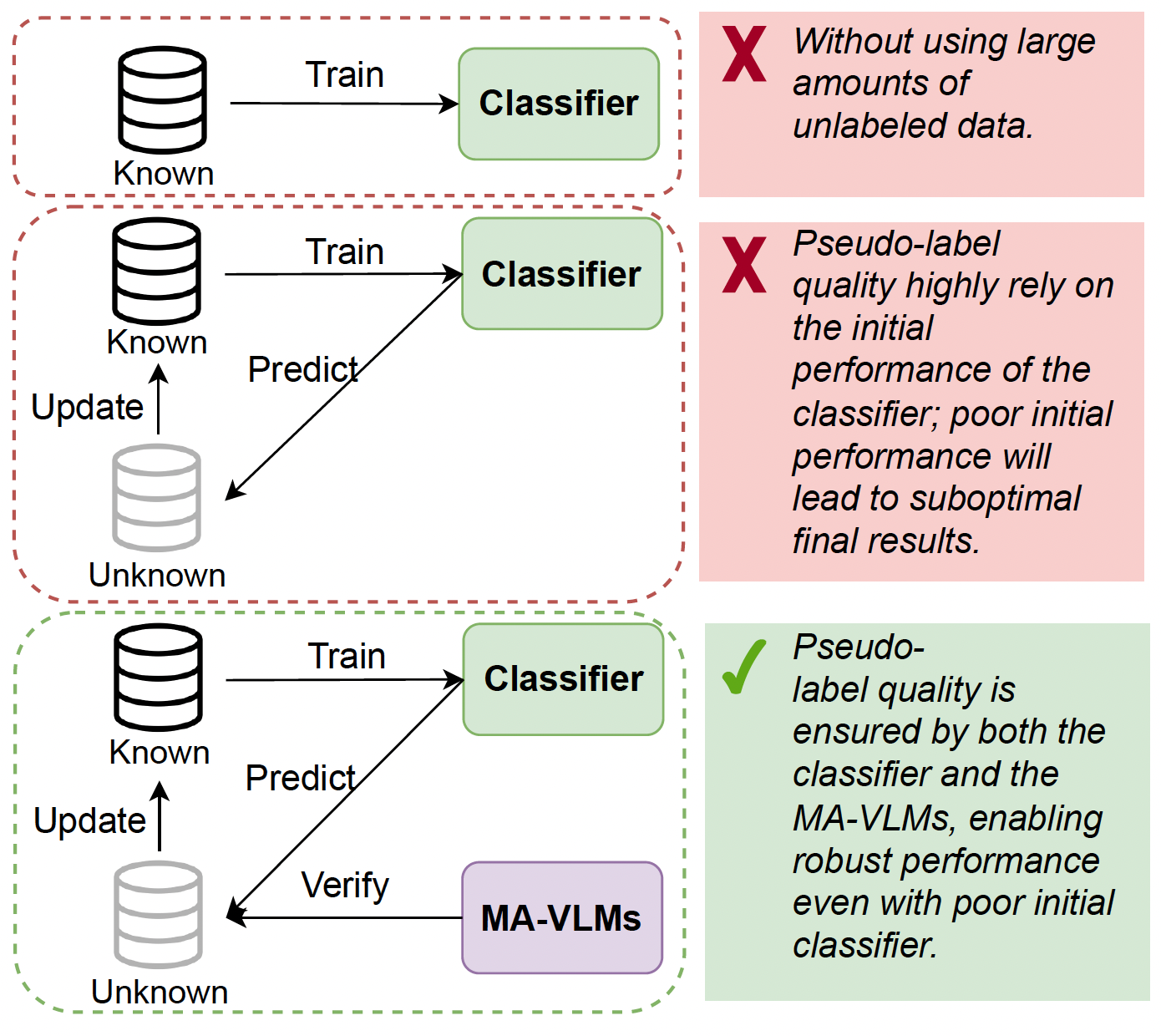}
  \caption{Comparison of our approach (bottom) with supervised-only (top) and traditional self-training (middle).}
  \label{fig:overall_diagram}
\end{figure}

To tackle offensive content in low-resource settings, researchers have explored several strategies, each with limitations. LLM prompting delivers strong few-shot results but is impractical for large-scale use due to high costs and latency. Data augmentation can help to balance datasets, but is limited to text and often lacks semantic diversity due to extreme label scarcity \cite{rizos2019augment, madukwe2022token, cao2020hategan}. Transfer learning from high-resource domains aids generalization but still requires a moderate amount of target-domain labels \cite{plaza2021multitask, wang2025crossmodal, deoliveira2024hate}. Self-training bootstraps from unlabeled data but is highly error-prone when initial models are trained with minimal supervision \cite{alsafari2021semi}. Moreover, most methods assume unimodal inputs and overlook challenges like label ambiguity and fairness in pseudo-labeling, both critical in real-world application.


We propose a novel self-training framework that addresses these limitations by combining a lightweight classifier with Multi-Agent Vision-Language Models (\textsf{MA-VLMs}) to guide the pseudo-labeling process (Figure~\ref{fig:overall_diagram}, bottom). The classifier begins training from a small labeled set and iteratively incorporates high-confidence predictions over unlabeled data. Rather than relying solely on classifier’s output, we simulate a multiperspective moderation process where two VLM agents act as a moderator (with safety-first bias) and a user (defending free expression). These agents engage in a prompt-based negotiation on unlabeled samples. If all three agree on a label (positive or negative), the sample is assigned a pseudo-label and added to the training pool as an \textit{Agreed-Unknown} instance. If there is any disagreement between the classifier and either agent, the sample is retained as \textit{Disagreed-Unknown}, a category used to capture uncertain or contested content. This distinction enables the model to treat high-consensus and low-consensus samples differently, enhancing both fairness and robustness in the learning process. This multi-agent negotiation mimics the real-world tension between over-censorship and under-protection, promoting greater fairness and reliability in label propagation.

We also propose a novel Positive-Negative-Unlabeled (PNU) loss to leverage labeled, agreed, and disagreed data. Building on PU learning framework \cite{elkan2008learning,sakai2017semi}, our approach combines: (i) standard positive-negative supervision for labeled data, (ii) soft supervision for \textit{Agreed-Unknown} data to mitigate overfitting to pseudo-labels, and (iii) a PU/NU-style loss for \textit{Disagreed-Unknown} data, modulated by a dataset-specific weighting factor $\gamma$. This hybrid formulation enables robust learning under label noise and allows the model to benefit from ambiguous samples often discarded in traditional pipelines.


We evaluate our framework on four benchmark datasets across multimodal and text-only settings, under limited supervision with varying number of labeled examples. Our method outperforms supervised baselines and rivals large VLMs using a lightweight classifier without VLM fine-tuning. Ablation studies validate the contributions of both the MA-VLM prompting format and the PNU loss.


We summarize our key contributions as follows: (i) We propose a novel self-training framework guided by \textsf{MA-VLMs}, designed for scalable and fair offensive content detection in low-resource settings. (ii) We introduce a socially-informed prompting framework that simulates real-world tensions between moderator and user perspectives, enabling more reliable pseudo-label generation. (iii) We propose a PNU-based loss that jointly leverages labeled data, high-confidence pseudo-labels, and disagreement-prone instances via confidence weighting. (iv) We demonstrate strong empirical performance on four benchmark datasets, showing that our method achieves robust results with as 
few as 50 labeled samples and outperforms larger models in several settings.

%% file: related.tex
\subsection{Low-Resource Offensive Content Detection}
Detecting offensive content requires substantial annotated data due to the implicit meanings, cultural context, and subjectivity involved~\cite{lu-etal-2025-llm, xiao-etal-2024-toxicloakcn, hu2025toxicity}. However, such data is scarce as offensive instances are often removed by moderation systems, and annotation is costly and complex. In contrast, unlabeled user-generated data is abundant and easily accessible. 

When labeled data are scarce, LLMs are a common fallback, leveraging zero- and few-shot capabilities to perform competitively with minimal supervision~\cite{chiu2021gpt3}. Their reasoning also enables explainability via rationale generation~\cite{nirmal2024interpretable,wang2023evaluating,hee2025demystifying}, though high inference costs limit scalability for real-time moderation\cite{hee2025contrastive,cao2024modularized,cao2023procap}.

With moderate supervision, techniques like data augmentation and transfer learning are often used. Augmentation addresses class imbalance through synonym replacement, token warping, or class-conditional generation~\cite{rizos2019augment, madukwe2022token, cao2020hategan}, but is typically text based and prone to overfitting. Transfer learning adapts models from related domains~\cite{plaza2021multitask,wang2025crossmodal, deoliveira2024hate, 10.1145/3746027.3758289, hee-etal-2024-bridging}, but demands non-trivial labeled data and struggles under extreme label scarcity.

Self-training (Figure~\ref{fig:overall_diagram}, middle) is a widely used approach in low-resource settings, where a classifier trained on limited labeled data is iteratively refined using pseudo-labels on unlabeled data. It has shown success in tasks such as text/image classification, NER, and speech recognition~\cite{amini2025selftraining}. However, its effectiveness depends on the quality of pseudo-labels. To improve this, methods such as confidence thresholding~\cite{tur2005combining,lee2013pseudo}, proportion-based selection~\cite{zou2018unsupervised,cascante2021curriculum}, adaptive thresholding~\cite{wang2023freematch,chen2023softmatch}, and dual-classifier frameworks~\cite{karamanolakis2021selftraining,chen2021semi} have been proposed. 
Some works have applied self-training to VLMs for VQA, image captioning \cite{yang2023self}, and object detection \cite{zhao2024taming, xu2023dst}. In offensive content detection, \cite{alsafari2021semi} uses 9K labeled and 5M unlabeled tweets but struggles in low-resource settings due to pseudo-label bias. Additionally, traditional self-training often overlooks the ambiguity and subjectivity in social tasks, limiting fairness and interpretability.

To overcome the limitations of self-training under scarce labeled data, we draw inspiration from knowledge distillation that transfer general LLM capabilities to task-specific models \cite{hsieh2023distilling}. Specifically, we propose a self-training framework (Figure~\ref{fig:overall_diagram}, bottom) that integrates a lightweight classifier with MA-VLMs. The classifier offers efficient inference, while MA-VLMs validate predictions through a negotiation process. Only samples where both agents and the classifier agree are assigned pseudo-labels and included as Positive or Negative \textit{Agreed-Unknown}. Disagreements are retained as \textit{Disagreed-Unknown}, allowing selective use of uncertain examples during training.

\subsection{LLMs Based Offensive Content Detection}
LLMs are increasingly applied to offensive content detection using prompting strategies such as Zero-Shot, Few-Shot, and Chain-of-Thought (CoT)~\cite{hee-etal-2024-recent}. Few-Shot prompting, which incorporates demonstration examples, notably improves GPT-3’s performance over Zero-Shot~\cite{chiu2021gpt3,10.1145/3664647.3681521}. Instruction-tuned LLMs have also enhanced Zero-Shot performance for these tasks~\cite{plaza2023respectful}. CoT prompting has been explored to support reasoning in social tasks like hate speech detection~\cite{gupta2024hate,wang2023evaluating}, though its effectiveness remains limited beyond structured domains such as mathematics~\cite{sprague2024tocot}. Recent studies underscore the potential of multi-agent LLM systems for complex decision-making and simulation~\cite{guo2024multiagent}.A related application employs multi-agent debate between LLMs for adversarial attack defense~\cite{chern2024debate}. However, most existing LLM-based approaches to offensive content detection rely on single models or homogeneous agents, reflecting a narrow viewpoint and neglecting the nuanced trade-off between safety enforcement and free expression.

To this end, we propose \textsf{MA-VLMs}, a multi-agent vision-language framework that simulates diverse social perspectives using two distinct VLMs. Unlike prior single-agent approaches, \textsf{MA-VLMs} introduces a agent-based negotiation mechanism to assess offensive content accounting for both safety enforcement and user expression.

\subsection{Positive-Unlabeled (PU) Learning}
PU learning tackles binary classification using a small set of labeled positives and a large pool of unlabeled data. One strategy applies self-training by initially treating all unlabeled data as negative and iteratively adding confident negatives for retraining~\cite{yu2005new, fusilier2015detecting}. Another uses risk estimation; since binary classification requires both positive and negative risks, \citet{elkan2008learning} proposed estimating the negative risk using only positive and unlabeled data, with \citet{kiryo2017positive} later introducing a non-negative estimator for greater stability. Traditional PU learning assumes a known positive prior, though some methods estimate it via mixture proportion~\cite{garg2021mixture}. Multi-class extensions~\cite{shu2020learning} and recent approaches combining self-training with risk estimation~\cite{chen2020selfpu} have also been proposed.


This concept extends to NU learning, which assumes only negative labels are available.
\citet{sakai2017semi} further proposed PNU learning, unifying labeled positive, labeled negative, and unlabeled data with a tunable parameter balancing the unlabeled contribution. 
Although PU learning has been applied in bioinformatics, business analytics, security, and signal processing~\cite{jaskie2019pu}, its use in subjective or social tasks, such as detection of offensive content, where ambiguity and disagreement among annotators are common, remains limited.

To address this gap, we introduce a novel PNU loss inspired by \citet{sakai2017semi}, adapted for use in our self-training framework. Our loss formulation leverages all available supervision: labeled, confidently pseudo-labeled (\textit{Agreed-Unknown}), and ambiguous (\textit{Disagreed-Unknown}) instances. This design enables effective training in low-resource settings by maximizing the utility of unlabeled data while mitigating the risk of label noise.



%% file: model.tex
\begin{figure*}[t]
  \centering
  \small
  \includegraphics[width=0.94\textwidth]{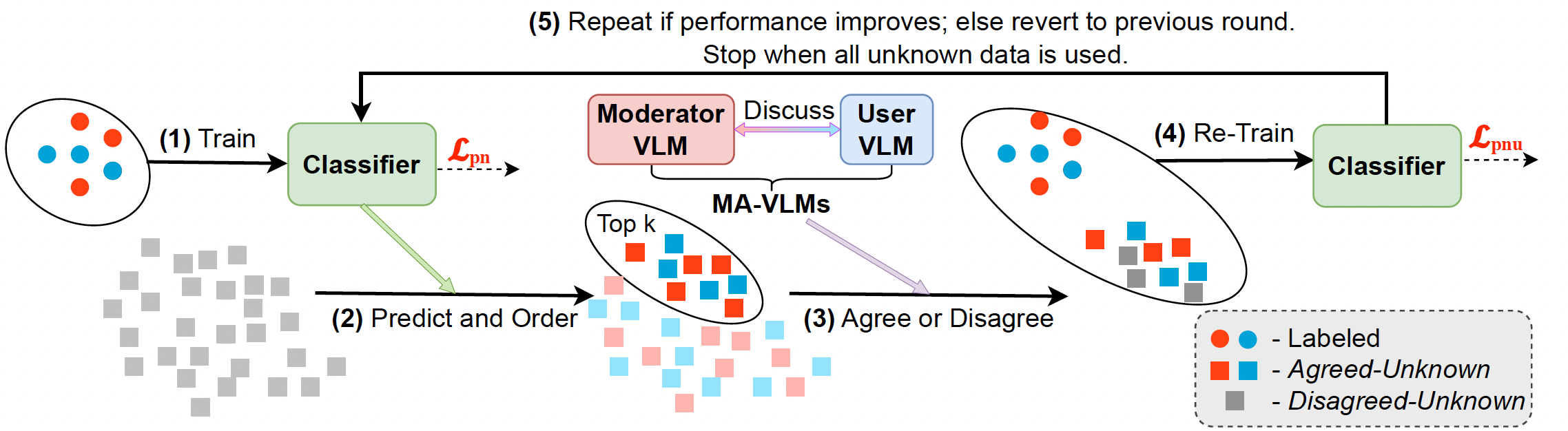}
  \caption{\textsf{MA-VLMs} guided self-training pipeline using PNU loss.}
  \label{fig:self_training_pipeline}
\end{figure*}

\begin{figure}[t]
  \centering
  \small
  \includegraphics[width=0.47\textwidth]{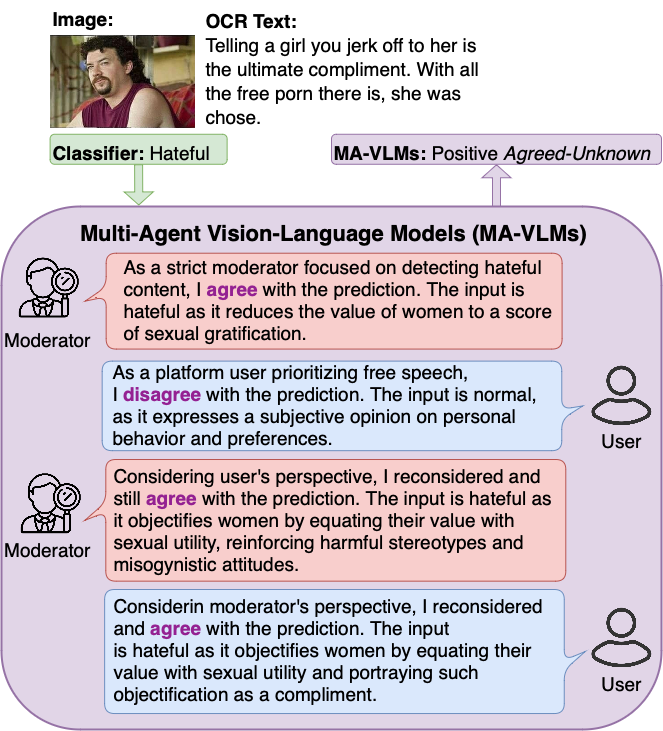}
  \caption{\textsf{MA-VLMs} with hate meme detection example.}
  \label{fig:MA_VLMs}
\end{figure}

Detecting offensive content spans diverse subtasks with varying labels, languages, and modalities, reflecting its social complexity and contextual nuance. Some subtasks face a scarcity of labeled data, while unlabeled data is plentiful. This creates a low-resource setting where only a small subset $n \ll N$ of the available $N$ samples is labeled. Our work focuses on leveraging this limited labeled data, with the help of abundant unlabeled data, to enhance detection performance. Figure~\ref{fig:self_training_pipeline} illustrates the framework employing self-training of a lightweight classifier with pseudo-labels generated by the classifier and \textsf{MA-VLMs}. Classifier retraining employs a PNU loss to integrate all data types. Details on the self-training pipeline, \textsf{MA-VLMs}, and PNU loss follow.

\subsection{\textsf{MA-VLMs} Guided Self-Training Pipeline}
In low-resource settings, classifiers often produce unreliable pseudo-labels, causing error propagation during retraining. To mitigate this, we propose the \textsf{MA-VLMs} guided self-training pipeline (Figure~\ref{fig:self_training_pipeline}), which leverages VLMs' offensive content understanding to verify classifier-generated pseudo-labels. The pipeline consists of five steps:

\begin{enumerate}
   \item \textbf{Train:} The classifier is trained on the $n$ labeled samples.
    
    \item \textbf{Predict and Order:} The trained classifier predicts unlabeled samples, ranking them by confidence scores.

    \item \textbf{Agree or Disagree:} The top $k$ confident predictions are verified by \textsf{MA-VLMs}. Agreed cases are pseudo-labeled as Positive or Negative \textit{Agreed-Unknown}; disagreements remain unlabeled as \textit{Disagreed-Unknown}.

    \item \textbf{Retrain:} Both agreed and disagreed samples are removed from the unlabeled pool and added to the training set. The classifier is retrained using a novel PNU loss $\mathcal{L}_{pnu}$ that integrates labeled, \textit{Agreed-Unknown}, and \textit{Disagreed-Unknown} samples.
    
    \item \textbf{Validation Check:} If development set performance improves after retraining, training advances to the next round; otherwise, the classifier reverts to its prior state, and the current top $k$ pseudo-labels are discarded.
\end{enumerate}

The procedure ends once all unlabeled samples are utilized. By verifying classifier predictions with \textsf{MA-VLMs} and selectively retaining only the top $k$ pseudo-labeled samples that improve performance, our pipeline achieves strong results even with very limited labeled data (e.g., $n=50$).

\subsection{Multi-Agent Vision-Language-Models (\textsf{MA-VLMs})}
Current large language model applications in offensive content detection typically adopt a single-agent moderator perspective. In contrast, real-world moderation balances moderators’ focus on safety with users’ desire for free expression. To better capture this dynamic, we propose the \textsf{MA-VLMs} prompt format (Figure~\ref{fig:MA_VLMs}), comprising two VLMs with contrasting personas: a strict moderator and a lenient user. Each agent first provides an initial decision with rationale, then reviews the other’s judgment before issuing a final decision. Inputs are labeled \textit{Agreed-Unknown} (Positive/Negative) only if both agents agree with the classifier; otherwise, they are \textit{Disagreed-Unknown}.

In the example shown, a meme initially predicted as hateful by the classifier is first disagreed with by the user agent. After reviewing the rationale of the moderator, the user reviews its judgement, uncovering a nuanced insight previously unrecognized: The meme frames the degradation of women as a compliment, masking its hateful intent. This highlights \textsf{MA-VLMs}’ ability to detect implicit hate by combining moderator and user perspectives.

\subsection{Preliminary of Positive-Unlabeled (PU) Learning}

The standard PN learning loss function is:

\begin{equation}
\label{eq:pn_loss}
\mathcal{L}_{\text{pn}} = \pi_p \mathcal{L}_p^{y_p} + \pi_n \mathcal{L}_n^{y_n}, \quad
\left\{
\begin{array}{l}
\mathcal{L}_p^{y_p} = \frac{1}{n^p} \sum\limits_{i=1}^{n^p} \ell(g(x_i^p), {y_p}), \\
\mathcal{L}_n^{y_n} = \frac{1}{n^n} \sum\limits_{i=1}^{n^n} \ell(g(x_i^n), {y_n})
\end{array}
\right.
\end{equation}
where $\pi_p$ and $\pi_n$ are the positive and negative class priors ($\pi_p + \pi_n = 1$); $\ell$ is the classification loss; $g(x)$ the model prediction; and $y_p$, $y_n$ the target labels for positive and negative samples $x_i^p$, $x_i^n$, with counts $n^p$ and $n^n$ respectively.

In PU learning, only positive and unlabeled data are available, rendering direct negative loss computation infeasible. To address this, PU learning \cite{elkan2008learning} approximates negative loss from positive and unlabeled samples. \cite{kiryo2017positive} further propose the non-negative PU loss $\mathcal{L}_{\text{pu}}$ defined below (full derivation in Appendix 1).
\begin{equation}
\mathcal{L}_{\text{pu}} = \pi_p \mathcal{L}_p^{y_p} + \max(0, \mathcal{L}_u^{y_n} - \pi_p \mathcal{L}_p^{y_n}).
\end{equation}

This loss estimates negatives without negative data and is widely used in fields where positives are easier to obtain, such as bioinformatics, analytics, security, and signal processing~\cite{jaskie2019pu}.

\subsection{Positive-Negative-Unlabeled (PNU) Loss}
Our pipeline uses three data types: labeled, \textit{Agreed-Unknown}, and \textit{Disagreed-Unknown}. \textit{Agreed-Unknown} samples are model-generated without human verification; to mitigate overconfidence, we assign soft targets. \textit{Disagreed-Unknown} data, where the classifier and \textsf{MA-VLMs} disagree, remain unlabeled and are typically discarded in PN learning. Inspired by PNU learning \cite{sakai2017semi}, which extends PU learning by incorporating unknown data via a weighting parameter $\gamma$, we propose a customized PNU loss to include these samples. The final loss differentiates the three data types to enhance robustness:

\begin{equation}
\mathcal{L}_{\text{pnu}} = 
\begin{cases}
(1 - \gamma) \cdot (\mathcal{L}_{\text{pn}} + \mathcal{L}_{\text{soft-pn}}) + \gamma \cdot \mathcal{L}_{\text{pu}}, & \text{if } \gamma \geq 0 \\
(1 + \gamma) \cdot (\mathcal{L}_{\text{pn}} + \mathcal{L}_{\text{soft-pn}}) - \gamma \cdot \mathcal{L}_{\text{nu}}, & \text{if } \gamma < 0
\end{cases}
\end{equation}
where $\gamma \in [-1, 1]$ is dataset-dependent: $\gamma < 0$ yields NU learning (label-reversed PU), $\gamma = 0$ reduces to PN learning, and $\gamma > 0$ corresponds to PU learning. The magnitude of $\gamma$ controls the strength of PU or NU learning.
The PN loss $\mathcal{L}_{\text{pn}}$ follows Equation~\ref{eq:pn_loss}.

The soft-PN loss $\mathcal{L}_{\text{soft-pn}}$ uses \textit{Agreed-Unknown} samples with soft labels to capture uncertainty:
\begin{equation}
\mathcal{L}_{\text{soft-pn}} = \pi_p \cdot \mathcal{L}_{pu}^{\hat{y}_p} + \pi_n \cdot \mathcal{L}_{nu}^{\hat{y}_n}
\end{equation}
where $\hat{y}_p$ and $\hat{y}_n$ are soft targets for Positive and Negative \textit{Agreed-Unknown} samples $x_i^{pu}$ and $x_i^{nu}$, with counts $n^{pu}$ and $n^{nu}$, respectively.

Inspired by PU learning, we treat labeled positives and \textit{Agreed-Unknown} samples as positive data, and \textit{Disagreed-Unknown} samples as unlabeled, yielding the PU loss $\mathcal{L}_{\text{pu}}$:
\begin{equation}
\mathcal{L}_{\text{pu}} = \pi_p ( \mathcal{L}_p^{y_p} +\mathcal{L}_{pu}^{\hat{y}_p} ) + \max(0, \mathcal{L}_u^{y_n}-\pi_p \cdot (\mathcal{L}_p^{y_n}+\mathcal{L}_{pu}^{\hat{y}_n} ) )
\end{equation}
Given the availability of negative data in our setting, the PU loss $\mathcal{L}_{\text{pu}}$ complements the negative component of PN learning, while the NU loss $\mathcal{L}_{\text{nu}}$, formulated similarly, complementing the positive component.

By tailoring losses to each data type, we account for the lower reliability of \textit{Agreed-Unknown} samples and exploit the complementary role of \textit{Disagreed-Unknown} data, leading to a more comprehensive loss formulation.

%% file: experiment.tex
\begin{table*}[t]
\centering
\small
\renewcommand{\arraystretch}{1.2}
\setlength{\tabcolsep}{6pt}
\begin{tabular}{l|c|c|c|cc|cc|cc|cc}
\hline
\textbf{Model} & \textbf{\#Params} &  \textbf{Train} & \textbf{PL Model} 
& \multicolumn{2}{c|}{\textbf{FHM}} 
& \multicolumn{2}{c|}{\textbf{MAMI}} 
& \multicolumn{2}{c|}{\textbf{HSOL}} 
& \multicolumn{2}{c}{\textbf{Sent140}} \\
\cline{5-12}
& & & & \textbf{Acc} & \textbf{M-F1} & \textbf{Acc} & \textbf{M-F1} & \textbf{Acc} & \textbf{M-F1} & \textbf{Acc} & \textbf{M-F1} \\
\hline
\hline
Qwen7B & 7B & SupOnly & - & 70.78 & \underline{70.41} & \textbf{76.20} & \textbf{76.06} & 85.00 & 84.89 & \textbf{78.20} &  \textbf{78.19} \\
RGCL & 428M & SupOnly & - & 35.78 & 26.35 & 50.00 & 33.33 & 48.00 &  32.43 & 48.10 & 33.28 \\
CLIP & 428M & SupOnly & - & 64.11 & 59.24 & 63.50 & 62.18 & 85.30 & 85.30 & 64.40 & 64.22 \\
\hline
CLIP & 428M & SelfTrain & CLIP & 
\underline{72.11} & 70.00 & 68.70 & 67.03 & \underline{86.50} & \underline{86.48}  & 73.10 & 73.05 \\
CLIP & 428M & SelfTrain & Qwen72B & 65.33 & 65.22 & 69.00 & 67.42 & 81.10 & 81.06 & 75.70 & 75.57 \\
CLIP & 428M & SelfTrain & CLIP + Qwen72B & \textbf{74.22} & \textbf{72.68} & \underline{73.80} & \underline{73.49} & \textbf{86.70} & \textbf{86.69} & \underline{77.40} & \underline{77.11} \\
\hline
\end{tabular}
\caption{Comparison of models and training strategies across four datasets ($n = 100$). PL Model = pseudo-labeling model. Top values highlighted; second-best underlined.}
\label{table:overall_results}
\end{table*}

\begin{table*}[ht]
\centering
\small
\begin{tabular}{@{}p{8.3cm}|c|c|c|c@{}}
\hline
\textbf{Text} & \textbf{GT} & \textbf{CLIP} & \textbf{Qwen72B} & \textbf{CLIP + Qwen72B}  \\ \hline \hline
 do you know how to use oven! or dig up hitler so he can show you & \textit{hateful} & \textit{hateful} & \textit{hateful} & Positive \textit{Agreed-Unknown} \\ \hline

 kermit the frog definitely not a muslim & \textit{hateful} & \textit{hateful} & \textit{non-hateful} & \textit{Disagreed-Unknown}
 \\ \hline
jerk off to a girl is compliment, with free porn, she was chosen & \textit{non-hateful} & \textit{hateful} & \textit{hateful} & Positive \textit{Agreed-Unknown} \\ \hline
\end{tabular}
\caption{Example of pseudo-labeling on FHS dataset; only text is shown for clarity.}
\label{table:demo_example}
\end{table*}

\begin{figure}[t]
  \centering
  \includegraphics[width=0.47\textwidth]{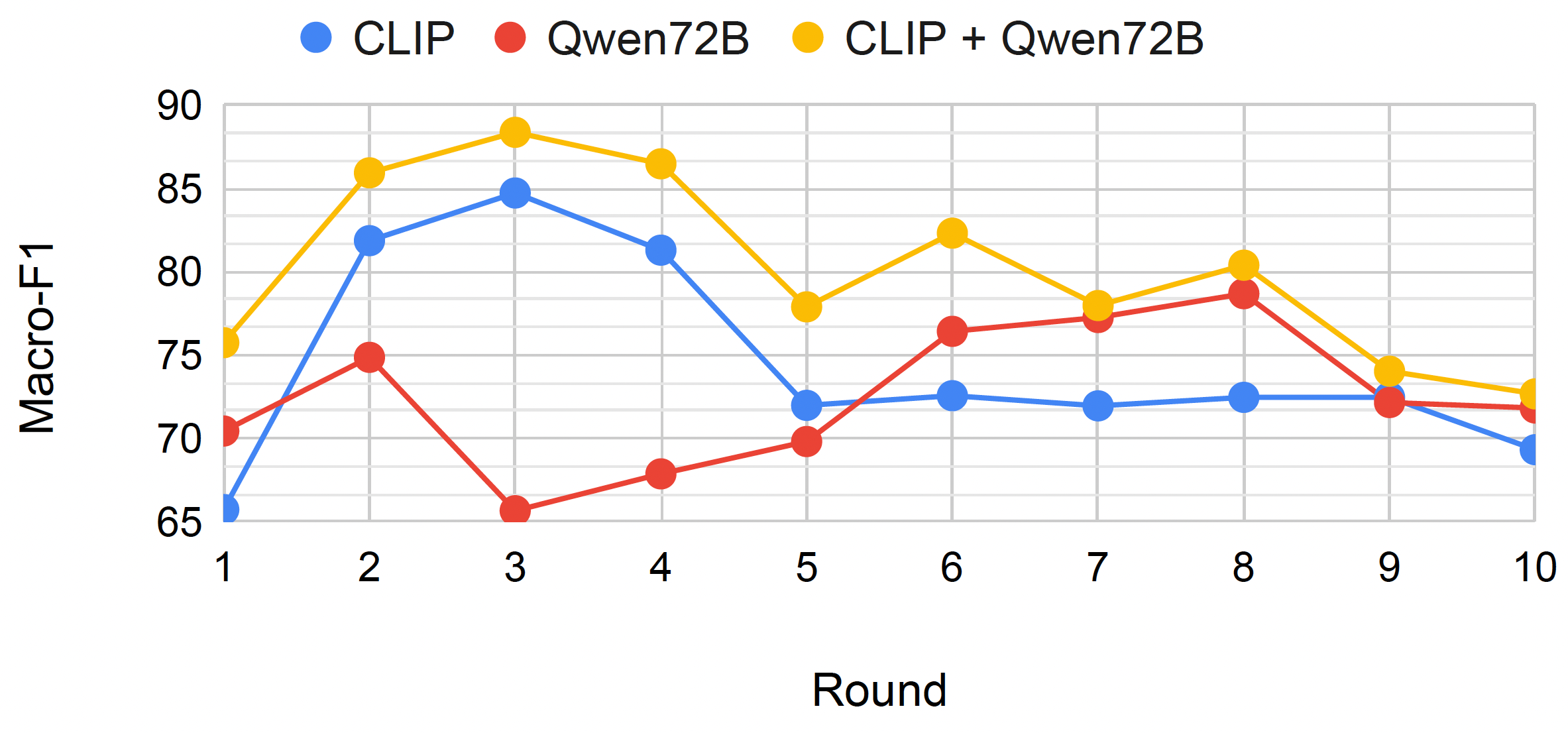}
  \caption{M-F1 of top $k$ pseudo-labeled samples per round on FHM ($n=100$); only first 10 rounds shown. GT = Ground Truth}
  \label{fig:PL_ACC}
\end{figure}



\subsection{Datasets}
To evaluate cross-modal and cross-task generalization, we experiment on four benchmarks: two multimodal and two text datasets. Three target offensive content: \textit{hate speech}, \textit{misogyny}, and \textit{general offensiveness}, while one assesses \textit{sentiment classification}, demonstrating robustness across related tasks. Dataset details follow:

\begin{itemize}
\item \textbf{Facebook Hateful Memes (FHM)}~\cite{kiela2020hatefulmemes}: A multimodal benchmark dataset of 10,000 Facebook memes labeled as \textit{hateful} or \textit{non-hateful}, designed for evaluating hate speech detection in memes.

\item \textbf{Multimedia Automatic Misogyny Identification (MAMI)}~\cite{fersini2022mami}: A multimodal dataset of 11,000 Instagram memes annotated for binary classification (Task A: \textit{misogynous} vs. \textit{non-misogynous}) and multi-label classification (Task B: \textit{shaming}, \textit{stereotype}, \textit{objectification}, \textit{violence}).

\item \textbf{Hate Speech and Offensive Language (HSOL)}~\cite{davidson2017hate}: A text dataset of 24,783 tweets labeled as \textit{hate speech}, \textit{offensive language}, or \textit{neither}, commonly used in online toxicity detection.

\item \textbf{Sentiment140 (Sent140)}~\cite{go2009twitter}: A large-scale dataset of 1.6 million tweets labeled for \textit{positive} or \textit{negative} sentiment, widely used in sentiment analysis tasks.
\end{itemize}

We frame all tasks as binary classification. In FHM and Sent140, \textit{hateful} and \textit{negative} samples are treated as Positive labels, while \textit{non-hateful} and \textit{positive} samples are Negative. For MAMI (Task A), \textit{misogynous} is Positive and \textit{non-misogynous} is Negative. In HSOL, \textit{hate speech }and \textit{offensive language} are grouped as Positive labels; \textit{neither} is Negative. All datasets were obtained from Kaggle. The FHM test set lacks labels and is excluded, resulting in 9,000 samples. Each multimodal dataset contains ~10,000 samples, while text-only datasets are larger. To ensure balance and manageable training time, we subsample 10,000 balanced examples from HSOL and Sent140 (see Appendix 2 for statistics).


\subsection{Models and Training Strategies}
To assess the effect of our self-training (SelfTrain) pipeline, we firstly compare it against the supervised-only (SupOnly) baseline. In the SupOnly setting, models are trained exclusively on $n$ labeled samples without access to unknown data. Three representative models are evaluated under this setup.

\begin{itemize}
\item \textbf{Qwen-2.5-VL-7B (Qwen7B)}~\cite{qwen2.5-VL}: Alibaba’s vision-language model, effective in fine-grained tasks such as VQA and captioning, is well-suited for multimodal offensive detection due to its robust multimodal and human intent understanding.
\item \textbf{RGCL}~\cite{mei2023improving}: A retrieval-guided contrastive framework using CLIP’s frozen embeddings to align inputs with hateful examples, achieving state-of-the-art hate meme detection.
\item \textbf{CLIP-Large (CLIP)}~\cite{radford2021learning}: A contrastive vision-language model projecting images and texts into a shared embedding space, has wide application in multimodal classification.
\end{itemize}

For SelfTrain (Figure~\ref{fig:self_training_pipeline}), we use CLIP-Large as the classifier and a frozen Qwen-2.5-VL-72B (Qwen72B) \cite{qwen2.5-VL} as the VLM in \textsf{MA-VLMs}. To assess the benefit of verifying classifier predictions with \textsf{MA-VLMs}, we compare three pseudo-labeling variants: (1) classifier-only, (2) \textsf{MA-VLMs}-only, and (3) their combination.

\subsection{Experiments Setting}
Each dataset is split into 80\% training, 10\% development, and 10\% test sets. Within the training set, only $n = 100$ samples are labeled; the rest remain unlabeled but are used for training. The development set is used for early stopping and the retention or removal of top $k$ pseudo-labels after each round, while the test set is reserved for final evaluation.

Qwen7B is fine-tuned using LoRA. Both RGCL and our method use pretrained CLIP-Large: RGCL extracts features to train a separate classifier, while our model fine-tunes CLIP-Large with a one-layer MLP. All models are trained for 10 epochs, with the best epoch selected based on development performance. Evaluation uses Accuracy (Acc) and Macro-F1 (M-F1), with M-F1 as the primary metric. We set $k = 500$ to balance performance and efficiency. For PNU loss, cross-entropy is used as $\mathcal{L}$, with $y_p = 1$, $y_n = 0$, and soft targets $\hat{y}_p = 0.67$, $\hat{y}_n = 0.33$. The class prior $\pi_p$ is fixed at 0.5, as deviations introduced class bias and degraded performance (Appendix 3.1). Optimal $\gamma$ values are determined via ablation: $\gamma = 0.0$ for FHM and $\gamma = 0.1$ for others.

\subsection{Results}

Table~\ref{table:overall_results} shows the performance of three models under SupOnly and three pseudo-labeling variants under SelfTrain, using $n = 100$ labeled samples across four datasets.

SelfTrain CLIP consistently outperforms SupOnly CLIP across all datasets and pseudo-labeling variants, highlighting the value of leveraging unlabeled data. Among the pseudo-labeling variants, the combination of CLIP and Qwen72B achieves the highest performance, surpassing either model individually by over 1.59 M-F1 points, except on HSOL where CLIP-only yields comparable results. These results show that the complementary strengths of CLIP and Qwen72B, combined with disagreement-based filtering, improve pseudo-label quality and overall performance.

Overall, SelfTrain CLIP with CLIP + Qwen72B pseudo-labeling achieves the best results on FHM and HSOL datasets, and ranks second on MAMI and Sent140 datasets, closely trailing the much larger SupOnly Qwen7B while consistently outperforming SupOnly RGCL. SupOnly Qwen7B's strong performance is attributed to its scale and extensive pretraining, though its large size limits its broader application. Although RGCL excels in high-resource settings (Table~\ref{table:known_data_n}), it struggles in low-resource scenarios due to its reliance on aligning inputs to labeled hateful samples, which hinders convergence under limited labeled data. In contrast, our SelfTrain CLIP with CLIP + Qwen72B pseudo-labeling is more effective in low-resource settings.

\subsection{Pseudo Label Analysis}
The strong performance of SelfTrain CLIP with CLIP + Qwen72B pseudo-labeling is primarily due to large-model verification and disagreement-based filtering, which improve pseudo-label quality. As shown in Figure~\ref{fig:PL_ACC}, the CLIP + Qwen72B setup consistently yields higher M-F1 for pseudo-labeled samples, particularly in early rounds. Performance declines as confident samples are exhausted, leaving ambiguous cases. This trend supports our strategy of confidence-based top-$k$ selection to ensure label quality.

Table~\ref{table:demo_example} shows pseudo-labeling examples from the FHM dataset (additional examples in Appendix 4). When CLIP and Qwen72B agree, inputs are typically clearly hateful, even if mislabeled in the ground truth (e.g., the last example), likely mislabeled due to annotator bias, which is a common issue across datasets. Disagreements reflect ambiguity, as in the middle example, where the hateful label is subjective and context-dependent. These observations support our strategy of assigning agreed samples as positive/negative and treating disagreed ones as unlabeled.

%% file: ablation.tex
\begin{table}[t]
\centering
\small
\renewcommand{\arraystretch}{1.2}
\setlength{\tabcolsep}{6pt}
\begin{tabular}{@{}c|c|c|cc|cc@{}}
\hline
\textbf{n} & \textbf{Model} & \textbf{Train} 
& \multicolumn{2}{c}{\textbf{FHM}} 
& \multicolumn{2}{c}{\textbf{MAMI}} \\
\cline{4-7}
& & & \textbf{Acc} & \textbf{M-F1} & \textbf{Acc} & \textbf{M-F1}  \\
\hline
\hline
50 & Qwen7B & SupOlny & 64.22 & 39.11 & 73.20 & 73.00 \\
50 & CLIP & SupOlny & 63.89 & 48.76 & 59.50 & 56.03  \\
50 & CLIP & SelfTrain & 73.00 & 71.27  & 71.40 & 71.40 \\
\hline
100 & Qwen7B & SupOlny & 70.78 & 70.41 & \textbf{76.20} & \underline{76.06} \\
100 & CLIP & SupOlny & 64.11 & 59.24 &  63.50 & 62.18\\
100 & CLIP & SelfTrain & 74.22 & 72.68 &  73.80 & 73.49 \\
\hline
250 & Qwen7B & SupOlny & \underline{77.67} & \underline{75.88} & \textbf{76.20} & \textbf{76.18}\\
250 & CLIP & SupOlny & 73.56 & 69.67 & 65.70 & 64.67 \\
250 & CLIP & SelfTrain & 74.89 & 72.97 & 75.00 & 74.79 \\
\hline
full & Qwen7B & SupOlny & \textbf{78.67} & \textbf{76.00} & 71.80 & 70.20 \\
full & RGCL & SupOlny & 78.22 & 76.17 & 76.10 & 75.34 \\
full & CLIP & SupOlny & 76.89 & 75.02 & 72.00 & 71.06 \\
\hline
\end{tabular}
\caption{Comparison of known data sizes ($n$ = 50 to full) on two multimodal datasets. full = all training data used. Top values highlighted; second-best underlined. }
\label{table:known_data_n}
\end{table}

\begin{table}[t]
\centering
\small
\renewcommand{\arraystretch}{1.2}
\setlength{\tabcolsep}{6pt}
\begin{tabular}{@{}c|c|c|cc|cc@{}}
\hline
\textbf{PF} &
\textbf{Model} & 
\textbf{Train} 
& \multicolumn{2}{c}{\textbf{FHM}} 
& \multicolumn{2}{c}{\textbf{MAMI}} \\
\cline{4-7}
& & & \textbf{Acc} & \textbf{M-F1} & \textbf{Acc} & \textbf{M-F1}  \\
\hline \hline
ZS & Qwen72B & No & 74.89 & \underline{74.46} & 79.40 & 79.17 \\
FS & Qwen72B & No & 71.22 & 71.09 & 75.80 & 75.08 \\
CoT & Qwen72B & No & \textbf{75.33} & 74.43 & 78.30 & 78.28 \\
MA & Qwen72B & No &  \underline{75.00} & \textbf{74.62} & \textbf{81.70} & \textbf{81.64} \\
\hline
\end{tabular}
\caption{Comparison of prompt formats on two multimodal datasets. PF = Prompt Format; ZS = Zero-Shot; FS = Few-Shot; CoT = Chain-of-Thought; MA = MA-VLMs. Top values highlighted; second-best underlined.}
\label{table:MA_VLM_discuss_results}
\end{table}


\begin{table}[t]
\centering
\small
\renewcommand{\arraystretch}{1.2}
\setlength{\tabcolsep}{6pt}
\begin{tabular}{@{}c|c|c|cc|cc@{}}
\hline
\textbf{$\gamma$} 
& \textbf{Model} 
& \textbf{Train} 
& \multicolumn{2}{c}{\textbf{FHM}} 
& \multicolumn{2}{c}{\textbf{MAMI}}\\
\cline{4-7}
& & & \textbf{Acc} & \textbf{M-F1} & \textbf{Acc} & \textbf{M-F1} \\
\hline
\hline
- 0.1 & CLIP & SelfTrain & 69.11 & 68.35 & 64.10 & 59.98 \\
0.0 & CLIP & SelfTrain & \underline{74.22} & \textbf{72.68} & 70.10 & 68.84 
\\
0.1 & CLIP & SelfTrain & 73.22 & 71.50 & \textbf{73.80} & \textbf{73.49}  \\
0.2 & CLIP & SelfTrain & \textbf{74.44} & \underline{71.79} & \underline{72.60} & \underline{72.42} \\
\hline
\end{tabular}
\caption{Comparison of $\gamma$ values $[-0.1, 0.0, 0.1, 0.2]$ in PNU loss on two multimodal datasets ($n=100$). Top values highlighted; second-best underlined.}
\label{table:theta}
\end{table}

\subsection{Number of Known Data ($n$)}
To evaluate the robustness of our self-training pipeline across varying labeled data sizes, Table~\ref{table:known_data_n} presents the performance of four models: Qwen7B, CLIP (SupOnly), RGCL, and CLIP (SelfTrain) with CLIP + Qwen72B pseudo-labeling, on two multimodal datasets for $n = 50, 100, 250,$ and full (all labeled). RGCL results for low-resource settings are omitted due to convergence failure (full results in Appendix 3.2). 
In extreme low-resource settings ($n = 50$), SupOnly models struggle with scarce labels, whereas SelfTrain maintains M-F1 scores above 70, evidencing its effectiveness. In moderately low-resource settings (e.g., $n=250$), our model underperforms Qwen7B, likely due to Qwen7B’s greater capacity for complex tasks.
In the high-resource setting, performance of Qwen7B and CLIP on the MAMI dataset unexpectedly declines from $n = 50$ to full, likely due to human annotation errors discussed in the pseudo-label analysis, which limit gains from additional data. RGCL's superior performance in the full setting demonstrates its robustness to these errors.

\subsection{Prompt Format}
Since our pseudo-labeling depends on frozen VLMs (Qwen72B), prompt format selection is crucial. Table~\ref{table:MA_VLM_discuss_results} reports the performance of four prompting strategies: Zero-Shot, Few-Shot, CoT, and our \textsf{MA-VLMs}, on two multimodal datasets (full results in Appendix 3.3). Few-Shot underperforms Zero-Shot, likely due to bias from randomly selected or unrepresentative examples. CoT offers no improvement, as its step-by-step reasoning is better suited to logical tasks than to social tasks requiring nuanced human judgment. Overall, \textsf{MA-VLMs} consistently outperform other formats, especially on MAMI, likely due to misogyny’s more ambiguous definition compared to hate speech. This highlights \textsf{MA-VLMs}' strength in handling label ambiguity through negotiation between agents.

\subsection{$\gamma$ Value in PNU Loss}
The parameter $\gamma$ governs the balance between PU/NU and PN learning. Table~\ref{table:theta} presents results for $\gamma \in {-0.1, 0.0, 0.1, 0.2}$ in the PNU loss used in SelfTrain CLIP with CLIP + Qwen72B pseudo-labeling on two multimodal datasets (full results in Appendix 3.4). Performance declines when $\gamma < 0$, indicating NU learning is ineffective in our setting. Similarly, large values ($\gamma > 0.1$) degrade performance, suggesting PU influence should be moderate. Optimal performance occurs for $\gamma$ between 0.0 and 0.1, depending on dataset characteristics. Offensive content datasets are often constructed using offensive lexicons, causing normal-labeled samples to retain offensive cues. In balanced datasets like MAMI and HSOL, this bias increases the proportion of Positive \textit{Agreed-Unknown} samples, making $\theta = 0.1$ more effective. In contrast, FHM’s imbalanced distribution (3:7 positive to negative) results in a more balanced \textit{Agreed-Unknown} set due to this bias, reducing the benefit of higher $\gamma$ values. For sentiment classification, such bias is absent, yielding comparable performance for $\theta = 0.0$ and $0.1$.

%% file: conclusion.tex
In summary, we propose a Multi-Agent Vision-Language Models (\textsf{MA-VLMs})-guided self-training framework with a tailored PNU loss for low-resource offensive content detection. Our experiments validate the effectiveness of key components: the dual-perspective prompt format (moderator and user) in \textsf{MA-VLMs}, verification of classifier predictions via \textsf{MA-VLMs}, and the PNU loss, which integrates labeled data with \textit{Agreed-Unknown} and \textit{Disagreed-Unknown} pseudo-labels. Moreover, the framework is both task- and model-agnostic, applicable beyond offensive content detection (e.g., sentiment classification), with no restrictions on classifier architecture or VLM inference.

Overall, the lightweight classifier trained via our self-training framework achieves performance comparable to much larger models, demonstrating robustness in extremely low-resource scenarios. This enables scalable, high-quality moderation of social media content in low-resource settings—such as under-resourced languages, modalities, or subtasks—requiring only tens to hundreds labeled samples, and highlights the potential social impact of our approach in promoting safer online environments.

%% file: Acknowledgement.tex
This research is supported in part by the National Research Foundation, Prime Minister’s Office, Singapore, and the Ministry of Digital Development and Information, under its Online Trust and Safety (OTS) Research Programme (Award Grant No. S24T2TS007), and  Ministry of Education, Singapore, under its Academic Research Fund (AcRF) Tier 2. Any opinions, findings, and conclusions or recommendations expressed in this material are those of the author(s) and do not reflect the views of the National Research Foundation, Prime Minister’s Office,  Singapore, or the Ministry of Digital Development and Information and the Ministry of Education, Singapore.